\documentclass[letterpaper, 10 pt, conference]{ieeeconf}  

\IEEEoverridecommandlockouts                              

\usepackage{amsmath, amsfonts, amssymb}
\usepackage{array}
\usepackage[caption=false]{subfig}
\usepackage{textcomp}
\usepackage{stfloats}
\usepackage{url}
\usepackage{verbatim}
\usepackage{graphicx}
\usepackage{stfloats}
\usepackage{color}
\usepackage{diagbox}
\usepackage{cite}
\usepackage{multirow}
\usepackage{caption}
\usepackage{makecell}
\usepackage[T1]{fontenc}
\usepackage[T1]{fontenc}
\usepackage{authblk}
\usepackage{subfig}
\usepackage{subfloat}
\usepackage{overpic}
\makeatletter
\let\NAT@parse\undefined
\makeatother
\usepackage[hidelinks]{hyperref}
\usepackage{marvosym}
\newcommand{\tabincell}[2]{\begin{tabular}{@{}#1@{}}#2\end{tabular}}
\usepackage{gensymb}

\title{\LARGE \bf
Revisiting Multi-modal 3D Semantic Segmentation in Real-world Autonomous Driving
}

\author{Feng Jiang$^{\dag,1}$, \and Chaoping Tu$^{\dag,2}$, \and Gang Zhang$^{\ast,\dag,2}$, \and Jun Li$^{2}$, \and Hanqing Huang$^{2}$, \and Junyu Lin$^{3}$, \and Di Feng$^{2}$, \and Jian Pu$^{\textsuperscript{\Letter}, 1}$ \and
\thanks{$\dag$ equal contribution}
\thanks{{\scriptsize \Letter} corresponding author, jianpu@fudan.edu.cn}
\thanks{$\ast$ zhanggang11021136@gmail.com}
\thanks{$^{1}$ISTBI, Fudan University}
\thanks{$^{2}$Mogo Auto Intelligence and Telematics Information Technology Co., Ltd }
\thanks{$^{3}$School of Computer Science, Fudan University}
}

\begin{document}

\maketitle
\thispagestyle{empty}
\pagestyle{empty}

\begin{abstract}

LiDAR and camera are two critical sensors for multi-modal 3D semantic segmentation and are supposed to be fused efficiently and robustly to promise safety in various real-world scenarios. However, existing multi-modal methods face two key challenges: 1) difficulty with efficient deployment and real-time execution; and 2) drastic performance degradation under weak calibration between LiDAR and cameras. To address these challenges, we propose CPGNet-LCF, a new multi-modal fusion framework extending the LiDAR-only CPGNet. CPGNet-LCF solves the first challenge by inheriting the easy deployment and real-time capabilities of CPGNet. For the second challenge, we introduce a novel weak calibration knowledge distillation strategy during training to improve the robustness against the weak calibration. CPGNet-LCF achieves state-of-the-art performance on the nuScenes and SemanticKITTI benchmarks. Remarkably, it can be easily deployed to run in 20\,ms per frame on a single Tesla V100 GPU using TensorRT TF16 mode. Furthermore, we benchmark performance over four weak calibration levels, demonstrating the robustness of our proposed approach.

\end{abstract}

\section{INTRODUCTION}
Scene understanding is very important for autonomous driving, and point cloud semantic segmentation is an important phase~\cite{guo2020deep}. Due to the sparsity of point clouds, the segmentation performance of distant objects is worse, which can be solved by introducing camera RGB information~\cite{li2023mseg3d, krispel2020fuseseg, zhao2023lif, zhuang2021perception}. Multi-modal semantic segmentation can simultaneously perceive the texture and geometric information of the scene. 
However, existing multi-modal methods are not ready for applications since the noteworthy actual problems that may happen in real-world scenarios are under-explored. In this paper, two prerequisites are introduced that the multi-modal methods should not only be easy-deployed and run in real-time but also demonstrate strong robustness against the weak calibration between LiDAR and cameras.


\begin{figure}[t]
\centering
\begin{overpic}
[width=1\linewidth, tics=5]{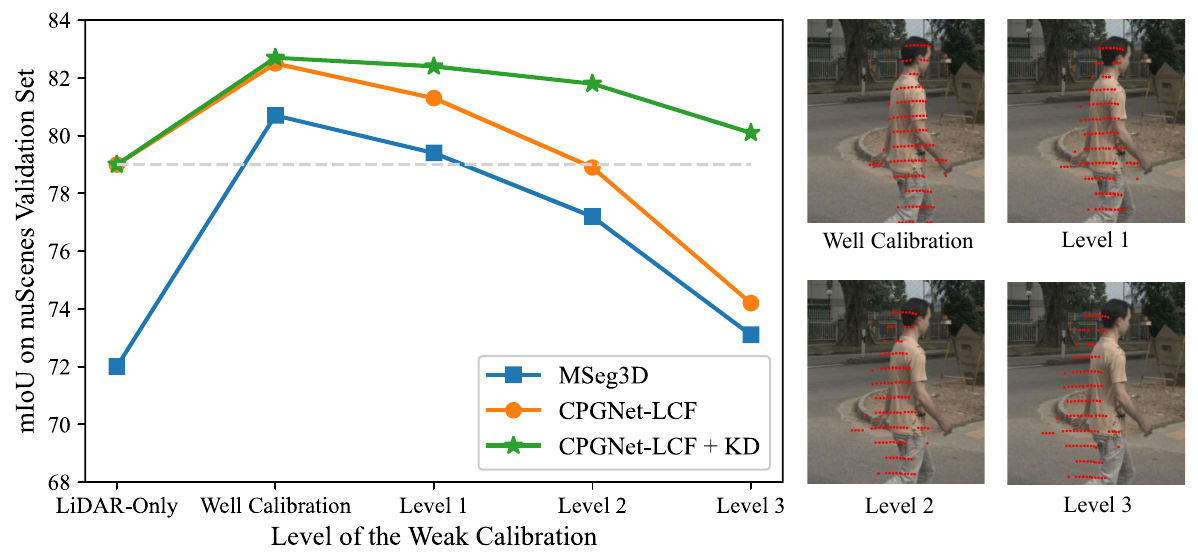} %
\end{overpic}
\caption{Impact of weak calibration on segmentation accuracy (mIoU). CPGNet-LCF and MSeg3D suffer from performance drops under weak calibration, even performing worse than the LiDAR-only ones. The proposed method mitigates these accuracy declines under weak calibration.}
\label{fig: first}
\end{figure}

Almost all existing multi-modal methods are built on the LiDAR backbones, resulting in the computational complexity primarily concentrated on them.
Currently, sparse voxel-based backbones~\cite{tang2020searching, zhu2021cylindrical, yan20222dpass} and 2D projection-based backbones~\cite{zhang2020polarnet, milioto2019rangenet++, cheng2022cenet, wu2019squeezesegv2} are receiving increasing attention due to their high accuracy or fast speed. 
The sparse voxel-based backbones are hard to use in real-world autonomous driving systems since they utilize the time-consuming 3D sparse convolution.
The 2D projection-based backbones can achieve faster inference speed by projecting the point clouds onto the 2D view, such as bird's-eye-view (BEV) and range view (RV), where the highly efficient 2D convolution network is applied.
Some previous works~\cite{li2022cpgnet, liong2020amvnet, qiu2022gfnet} further integrated the multiple 2D views for stronger representation, while keeping their efficiency. 
Another serious problem of the multi-modal methods~\cite{krispel2020fuseseg, zhuang2021perception, li2023mseg3d} is that they do not consider the weak calibration problem that inevitably occurs in practice. Fig.~\ref{fig: first} demonstrates how performance is affected by different weak calibration levels. Notably, at the Level 2, and 3 of weak calibration, the results are even worse than those of the LiDAR-only models. The projection deviations between the well and weak calibrations, as shown on the right side of Fig.~\ref{fig: first}, explain this performance degradation. 
Misalignment between LiDAR and cameras can introduce errors, guiding the multi-modal methods to learn wrong information.


To address the aforementioned challenges, we propose CPGNet-LCF, a LiDAR and Camera Fusion method that extends the easy-deployed CPGNet~\cite{li2022cpgnet} as the LiDAR backbone and integrates the lightweight 2D image semantic segmentation network STDC~\cite{fan2021rethinking} as the image backbone. 
Concretely, the image features are extracted by the STDC and further augment the LiDAR features by bilinear sampling according to the calibration matrices between LiDAR and cameras. Subsequently, the image-augmented LiDAR features are processed by the CPGNet to acquire the semantic segmentation results.

The proposed CPGNet-LCF is evaluated on nuScenes~\cite{caesar2020nuscenes} and SemanticKITTI~\cite{behley2019iccv} LiDAR semantic segmentation benchmarks and achieves the leading results (83.2 mIoU on the nuScenes leaderboard and 67.1 mIoU on SemanticKITTI validation set). CPGNet-LCF runs 63\,ms per frame with PyTorch and 20\,ms per frame with TensorRT TF16 inference mode on a single NVIDIA Tesla V100 GPU, which means that the proposed method is easy-deployed and real-time. Four levels of the weak calibration evaluation benchmarks based on nuScenes~\cite{caesar2020nuscenes} are established to evaluate the robustness of the model against the weak calibration. The proposed method outperforms the existing methods by a large margin while maintaining high performance even when the weak calibration level increases.

The main contributions of our work are as follows:
\begin{itemize}
  \item Existing multi-modal methods have difficulty in deployment, real-time execution, and robustness against weak calibration. We hope that researchers can pay more attention to the application-oriented issues.
  \item We extend the efficient CPGNet to the LiDAR and camera fusion, dubbed CPGNet-LCF. A novel weak calibration knowledge distillation strategy is adopted to alleviate the effects of the weak calibration between LiDAR and cameras.
  \item The proposed method achieves the SOTA results on two public datasets and shows remarkable robustness against weak calibration. Besides, it can be easily deployed on the TensorRT TF16 mode and runs 20\,ms per frame.
\end{itemize}

\section{RELATED WORK}

\subsection{LiDAR-Only Semantic Segmentation}
\label{sec: related lidar}
Due to the irregular and unordered properties of point clouds, methods for processing point clouds can be divided into point-based methods, sparse voxel-based methods, and 2D projection-based methods~\cite{lu2020deep}. 

The point-based methods directly process the original point cloud, which can retain more information, but requires nearest-neighbor searching to obtain surrounding information. PointNet~\cite{qi2017pointnet} proposed a simple but effective model, which obtained global features through the stack of MLPs. 
KPConv~\cite{thomas2019kpconv} proposed convolution operation on points, which learned weight kernel to mimic the traditional convolutions.
Sparse voxel-based methods usually convert point cloud into a dense or sparse representation. Huang \textit{et al.}~\cite{huang2016point} used fully-3D convolution to extract features for each voxel and Benjamin \textit{et at.}~\cite{graham2015sparse} further proposed sparse convolution to reduce the computational complexity. Cylinder3D~\cite{zhu2021cylindrical}, a classic voxel-based method, proposed to use cylindrical partition to fit the scanning mode of LiDAR. These methods are time-consuming due to the huge computation complexity and cannot be easily deployed to autonomous driving systems.

2D projection-based methods first project the point clouds into a certain 2D view, such as bird's-eye-view (BEV) and range view (RV). Squeezeseg~\cite{wu2018squeezeseg} introduced an efficient but effective backbone and RangeNet++~\cite{milioto2019rangenet++} proposed a post-processing method to solve the re-projection problem.
PolarNet~\cite{zhang2020polarnet} partitioned the points into grids with polar BEV coordinates and proposed ring convolution to better utilize the scanning of LiDAR. Furthermore, some works adopted a multi-view fusion approach to simultaneously benefit from BEV and RV representation. 
CPGNet~\cite{li2022cpgnet} proposed a Point-Grid fusion block, which fuses point, BEV and RV features in a cascade framework, which primarily consisted of 2D convolutions. 
In this paper, the CPGNet is adopted as the LiDAR backbone due to its effectiveness and efficiency.

\subsection{Multi-modal Fusion Semantic Segmentation}
\label{sec: related fusion}
The fusion of LiDAR and camera is an effective method to improve perception ability. The camera provides RGB and detailed information, while LiDAR provides more spatial and geometric information about the surrounding environment. Many previous works~\cite{krispel2020fuseseg, el2019rgb, zhuang2021perception, li2023mseg3d} have made significant contributions.
~\cite{krispel2020fuseseg} and ~\cite{el2019rgb} utilized the dense intrinsic distance representation and calibration information of LiDAR sensors to establish a point correspondence relationship between two input modalities. PMF~\cite{zhuang2021perception} proposed a LiDAR and camera fusion method based on perspective projection instead of spherical projection, which can exploit perceptual information from two modalities. MSeg3D~\cite{li2023mseg3d} proposed a multi-modal 3D semantic segmentation model that combined intra-modal feature extraction and inter-modal feature fusion, which to some extent alleviated the modal heterogeneity and also applied the asymmetric transformations to enhance the effectiveness of data augmentation.
LIF-Seg~\cite{zhao2023lif} predicted an offset to rectify the features of the two modalities, but the results were unstable and difficult to be widely applied.
The fusion of LiDAR and camera always needs an accurate calibration matrix from LiDAR to cameras, which is the cornerstone of the above models. Weak calibration apparently has a huge influence, but few works take this problem into consideration.  
\begin{figure*}[t]
\centering
\begin{overpic}
[width=1\linewidth, tics=5]{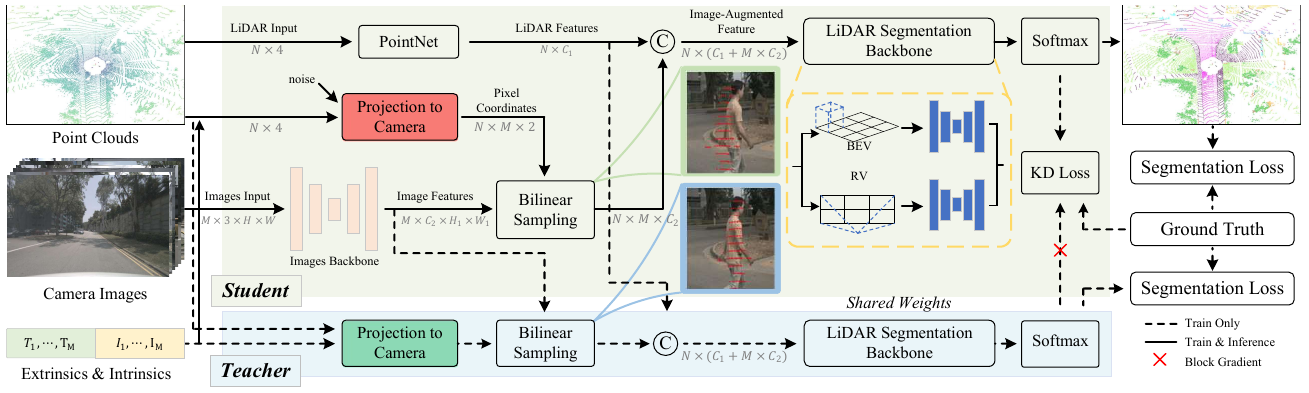} %
\end{overpic}
\caption{Illustration of the proposed LiDAR and camera fusion method CPGNet-LCF, which is extended from CPGNet~\cite{li2022cpgnet}. Projecting point clouds onto images is presented in detail, and visualization under well calibration (green part) and weak calibration (red part) are provided. Meanwhile, the proposed weak calibration distillation strategy transfers meaningful knowledge from the teacher (well calibration model) to the student (weak calibration model).}
\label{fig:fusion}
\end{figure*}

\section{METHOD}
\label{sec:method}
The real-world autonomous driving system needs an easy-deployed, real-time, and robust multi-modal method. 
To achieve the first two goals, we propose a LiDAR and camera fusion method, dubbed CPGNet-LCF, which is an extendsion of the easy-deployed CPGNet~\cite{li2022cpgnet}. For the last goal, a novel weak calibration knowledge distillation strategy is proposed.

In this section, we first illustrate the definitions of the semantic segmentation, calibration matrix, and weak calibration between LiDAR and cameras in Sec.~\ref{sec:pre}. Secondly, the details of the proposed CPGNet-LCF are described in Sec.~\ref{sec: overall}. Subsequently, the weak calibration knowledge distillation strategy is introduced in Sec.~\ref{sec: wckds} and the loss functions are listed in Sec.~\ref{sec:loss}. 

\subsection{Preliminary}
\label{sec:pre}
\subsubsection{Semantic Segmentation}
Given the input of point cloud $P\in \mathbb{R}^{N\times C_p}$, camera images $C=\{c_1, c_2, \cdots, c_M\}\in \mathbb{R}^{W\times H\times 3}$. $N, M$ denote the number of point clouds and the number of RGB images of each frame, and $C_p$ is the number of input channels of each LiDAR point (usually 4, including the XYZ coordinates and reflection intensity).
The objective of 3D semantic segmentation is to predict a label for each LiDAR point, and the number of semantic categories is $N_{cls}$.

\subsubsection{Calibration Matrix}
For LiDAR and camera fusion methods, the calibration matrix is used to align the point cloud 3D coordinates and image 2D pixels~\cite{zhang1999flexible, fremont2008extrinsic}.
The calibration matrices, also dubbed extrinsic matrices from LiDAR point cloud to each camera are represented as $T=\{T_1, T_2, \dots T_M\}\in \mathbb{R}^{4\times 4}$, where $M$ is the number of cameras. Each matrix is composed of rotation matrix $R_i\in \mathbb{R}^{3\times 3}$ and translation vector $t_i\in \mathbb{R}^{3\times 1}$, which presents the way to convert points from the LiDAR coordinate system to the camera coordinate system.
Intrinsic matrix of each camera is represented as $I=\{I_1, I_2, \dots I_M\}\in \mathbb{R}^{3\times 4}$, which reflects the attributes of camera. The point cloud that has been transformed into the camera coordinate system can be projected onto the image plane through the intrinsic matrix.

The point cloud can be projected onto the image $c_i$ through $T_i$ and $I_i$, which is the foundation for many previous works of multi-modal fusion. The formula of transformation is:
\begin{equation}
    \label{eq:project}
    \begin{aligned}
        \lambda
        \begin{bmatrix}
            u\\v\\1
        \end{bmatrix}
        =
        I_i
        \begin{bmatrix}
            R_i&t_i\\0&1\\
        \end{bmatrix}
        \begin{bmatrix}
            x\\y\\z\\1
        \end{bmatrix}
        = 
        I_i T_i
        \begin{bmatrix}
            x\\y\\z\\1
        \end{bmatrix},
    \end{aligned}
\end{equation}
which is an example of projecting a LiDAR point $\left(x, y, z\right)$ to the pixel $\left(u, v\right)$ of the $i\text{-th}$ camera image. 

\subsubsection{Weak Calibration}
\label{sec:weak}
In real-world autonomous driving scenarios, weak calibration inevitably occurs due to loose or deformed brackets securing LiDAR and the cameras. The calibration of LiDAR to the cameras involves angles (yaw, pitch, roll) and offsets ($t_x$, $t_y$, $t_z$). In practice, angles are more likely to change compared to offsets. \textbf{Therefore, in this paper, only the change in angles is discussed.} According to the projection relationship of Eq.~\ref{eq:project}, the weak calibration is equal to disturbing the extrinsic matrix $T_i$ into $T_i^e=T_i E_r$, which can be composed of the original extrinsic matrix $T_i$ and disturbing matrices $E_r=E_r^x E_r^y E_r^z$. The disturbing matrices $E_r^x$, $E_r^y$, $E_r^z$ only consider the rotation angles corresponding to the XYZ axes, respectively, where $r$ denotes the disturbing level of the weak calibration.
As shown in the right part of Fig.~\ref{fig: first}, the point cloud cannot be aligned with the image properly if the weak calibration occurs, which is fatal for the LiDAR and camera fusion methods. Apparently, the misaligned image features will introduce wrong information, and further degrade the performance of the multi-modal methods, even worse than that of the LiDAR-only counterpart, as shown in the left part of Fig.~\ref{fig: first}.


\subsection{Overall Framework}
\label{sec: overall}
The proposed CPGNet-LCF primarily consists of efficient and easy-deployed 2D convolution operators, which ensure the capabilities of easy deployment and fast inference speed. As shown in Fig.~\ref{fig:fusion}, it has three steps: 1) the image backbone is used to extract the meaningful texture features; 2) the LiDAR point features are augmented by bilinear-sampling the image features; 3) the image-augmented LiDAR features undergo the LiDAR segmentation backbone to acquire the final LiDAR semantic segmentation results. These three steps are introduced as the following.

Images have much higher resolutions than the sparse LiDAR point clouds and can provide rich texture information, which can assist the LiDAR in recognizing faraway and small objects. Generally, the 2D convolution network is used to process the images for efficiency.
STDC~\cite{fan2021rethinking}, used as the image backbone, proposes a lightweight backbone for segmentation tasks, which uses detailed guidance during training to keep more detailed information. The features $F\in\mathbb{R}^{M\times C_2\times H_{1}\times W_{1}}$ after the image backbone are used to augment the LiDAR point features. Note that the spatial size of $F$ is 1/8 of the input image size.

The image-augmented features fuse the semantic information from images and geometric information from LiDAR.
Specifically, the pixel coordinates $\left(u, v\right)$ corresponding to each LiDAR point can be obtained from the projection transformation relationship mentioned in Sec.~\ref{sec:pre}.
Bilinear sampling based on the image plane coordinates $\left(u, v\right)$ to obtain the corresponding features $f_I^i\in \mathbb{R}^{N\times C_2}$ of the $i$-th camera is
\begin{equation}
    f_I^i = \sum_{p=0}^1\sum_{q=0}^1\omega_{p,q}F_{i,\lfloor u \rfloor+p, \lfloor v \rfloor+q},
\end{equation}
where $\omega_{p, q} = (1-\left| (\lfloor u \rfloor+p) \right|)\cdot(1-\left| (\lfloor v \rfloor+q) \right|)$.
In general, autonomous vehicles have more than one camera, thus obtaining features of $M$ groups. However, the perspective field of each camera is limited, and features outside the perspective field are regarded as zeros when sampling.
Concatenating along the dimensions of the camera features $f_{I}^1, \cdots, f_{I}^M$ is used to obtain features $f_M \in \mathbb{R}^{N\times (M\times C_2)}$ for each LiDAR point.
The LiDAR point cloud features obtained from PointNet~\cite{qi2017pointnet} are $f_P\in \mathbb{R}^{N\times C_1}$. 
Finally, the sampling camera features $f_M$ and LiDAR point features $f_P$ are concatenated along the channel dimension to form the image-augmented features $f\in \mathbb{R}^{N\times (C_1+M\times C_2)}$.

LiDAR segmentation backbone serves as the basic part of CPGNet-LCF.
After obtaining the image-augmented features, it can be considered as the point cloud features with richer information, which makes it very convenient to use existing LiDAR segmentation models.
For both high accuracy and fast inference speed, a BEV and RV fusion framework CPGNet~\cite{li2022cpgnet} is adopted as the basic backbone. CPGNet primarily consists of efficient and easy-deployed 2D convolution operations instead of the time-consuming 3D sparse convolution operations. For more details on CPGNet~\cite{li2022cpgnet}, please refer to its original paper.


\subsection{Weak Calibration Knowledge Distillation Strategy}
\label{sec: wckds}
The weak calibration problem inevitably occurs in the real-world autonomous driving system, and attracts much less attention in the previous works. 
The definition and effect of the weak calibration have been illustrated in Sec.~\ref{sec:weak}. 
Therefore, we propose a novel weak calibration knowledge distillation strategy, which not only alleviates the effect of the weak calibration but also keeps the performance under the well calibration.

Before introducing the weak calibration knowledge distillation, we discuss a straightforward and simple idea, named weak calibration data augmentation. It treats the noise presented in the weak calibration matrix as a kind of data augmentation that mimics the weak calibration conditions during training. Consequently, it can help the model learn the inherent features of weak calibration and assist the multi-modal methods to adapt to the weak calibration.
In the experiment, it can be observed that the performance under weak calibration increases significantly but the performance under the well calibration drops. A possible reason is that this weak calibration data augmentation introduces much noise and misleads the training of the multi-modal methods.
Weak calibration knowledge distillation strategy is proposed to address the issue of performance degradation under the well calibration mentioned above.
Specifically, the model trained by the well calibration samples serves as the teacher model to guide the student model trained by weak calibration samples. 
It is obvious that the teacher model trained by the well calibration data, does not suffer from the performance degradation from the weak calibration data augmentation and thus has the ability to correctly guide the training of the student model. This knowledge distillation strategy can make the student model not only shows robustness against the weak calibration but also keeps performance under the well calibration. Note that the teacher model and student model share the same network parameters.

\subsection{Loss Function}
\label{sec:loss}
The overall loss function has two parts: the segmentation loss $\mathcal{L}_{\text {pc}}$ and weak calibration knowledge distillation loss $\mathcal{L}_{\text {wckd}}$.

The segmentation loss for point cloud $\mathcal{L}_{\text {pc}}$ follows the previous work~\cite{li2022cpgnet} and the detailed equation is given as 
$\mathcal{L}_{\text {pc}} = \mathcal{L}_{\text {wce}} + \mathcal{L}_{\text { Lovász }}$.
$\mathcal{L}_{\text {wce }}$ means weighted cross-entropy loss. 
$\mathcal{L}_{\text {Lovász}}$ is proposed by ~\cite{berman2018lovasz}, which directly optimizes the intersection-over-union (IoU) score. Given the probability $\hat{p}$ of the teacher model and the probability $p$ of the student model, the loss is calculated by:
\begin{equation}
    \mathcal{L}_{\text {wckd}} = -\hat{p}_c\log(p_c) - \lambda^2 \sum_{k\neq c}^{N_{cls}} \mathcal{N}(\hat{p}_k)\cdot \log(\mathcal{N}(p_k^\lambda)),
\end{equation}
where $c$ is the target category, $\lambda$ is the temperature of KD, and $\mathcal{N}$ means normalization functions.
Totally, the training objective of our method is 
\begin{equation}
\begin{aligned}
\mathcal{L}=\lambda_1\mathcal{L}_{\text {pc }}+\lambda_2 \mathcal{L}_\text{{wckd}}
\end{aligned}
\end{equation}
where $\lambda_1,\lambda_2$ are the balanced coefficients.

\section{EXPERIMENTS}

\subsection{Experimental Setup}

\textbf{nuScenes Dataset} is a public large-scale dataset for autonomous driving, which contains 1000 scenes of 20 seconds each~\cite{caesar2020nuscenes}. Specifically, 850 scenes of all scenes are used as the training and validation sets, while the remaining 150 are used as the test set. 
The dataset is annotated with keyframes selected at 2HZ. The point cloud of keyframes is labeled into 32 categories, including foreground categories such as \emph{Car} and \emph{Bus}, and background categories such as \emph{Sidewalk} and \emph{Manmade}. According to official requirements, we merge certain classes and only evaluate 16 classes.


\textbf{SemanticKITTI Dataset } is a large dataset for LiDAR point cloud semantic segmentation, which provides annotations for all points~\cite{behley2019iccv}. It has 22 point cloud sequences, each collected from a different scene. As recommended, we use 00 to 10 for training, 08 for validation, and 11 to 21 for testing. 
However, when collecting the dataset, only two front cameras were used, and the field of view was heavily overlapped. Therefore, we follow the setting of the PMF~\cite{zhuang2021perception}, which only uses point clouds in the overlapping area of the camera and LiDAR for evaluation.

\textbf{Metric} follows the previous work~\cite{li2023mseg3d}. We use the evaluation metric of the mean intersection-over-union (mIoU) over all classes, defined as $mIoU=\frac{1}{N_{cls}} \sum_{c=1}^{N_{cls}} \frac{T P_c}{T P_c+F P_c+F N_c}$, where $TP_{c}, FP_c, FN_c$ denote the number of true positive, false positives, and false negatives points of category $c$, respectively. The mIoU is the average of all classes.

\begin{table*}[t]
  \centering
  \scriptsize
  \caption{Quantitative comparisons on the nuScenes official leaderboard. The results are reported in terms of the per-class IoU, overall mIoU and latency of all methods. Here, L, C, and R respectively represent the use of LiDAR, camera, and Radar in the model. CPGNet-LCF achieves the best mIoU of 83.2 and the fastest inference speed 63\,ms per frame with PyTorch and 20\,ms per frame with TensorRT (noted with T).}
    \setlength{\tabcolsep}{1.2mm}{
    \begin{tabular}{c|c|cccccccccccccccc|cc}
    \hline
    & & & & & & & & & & & & & & & & & & &\\
    \multicolumn{1}{c|}{Methods} & {\rotatebox{90}{Modality}} & {\rotatebox{90}{Barrier}} & {\rotatebox{90}{Bicycle\hspace{-1cm}}} & {\rotatebox{90}{Bus\hspace{-1cm}}} & {\rotatebox{90}{Car\hspace{-1cm}}} & {\rotatebox{90}{C-vehicle\hspace{-1cm}}} & {\rotatebox{90}{Motorcycle\hspace{-1cm}}} & {\rotatebox{90}{Pedestrian\hspace{-1cm}}} & {\rotatebox{90}{Traffic-cone\hspace{-1cm}}} & {\rotatebox{90}{Trailer}} & \multicolumn{1}{l}{\rotatebox{90}{Truck}} & {\rotatebox{90}{D-surface}} & {\rotatebox{90}{Other-flat\hspace{-1cm}}} & {\rotatebox{90}{Sidewalk\hspace{-1cm}}} & {\rotatebox{90}{Terrain\hspace{-1cm}}} & {\rotatebox{90}{Manmade\hspace{-1cm}}} & {\rotatebox{90}{Vegetation\hspace{-1cm}}} & \multicolumn{1}{c}{{\rotatebox{90}{mIoU}}} & {\rotatebox{90}{{Latency(ms)}}} \\    \hline

    \multicolumn{1}{c|}{PolarNet~\cite{zhang2020polarnet}} & L & 72.2 & 16.8 & 77.0 & 86.5 & 51.1 & 69.7 & 64.8 & 54.1 & 69.7 & 63.5 & 96.6 & 67.1 & 77.7 & 72.1 & 87.1 & 84.5 & 69.4 & 62\\

    \multicolumn{1}{c|}{AMVNET~\cite{liong2020amvnet}} & L & 80.6 & 32.0 & 81.7 & 88.9 & 67.1 & 84.3 & 76.1 & 73.5 & 84.9 & 67.3 & 97.5 & 67.4 & 79.4 & 75.5 & 91.5 &  88.7 & 77.3 & - \\

    \multicolumn{1}{c|}{Cylinder3D++~\cite{zhu2021cylindrical}} & L & 82.8 & 33.9 & 84.3 & 89.4 & 69.6 & 79.4 & 77.3 & 73.4 & 84.6 & 69.4 & 97.7 & 70.2 & 80.3 & 75.5 & 90.4 & 87.6 & 77.9 & 142\\

    \multicolumn{1}{c|}{TVSN~\cite{hanyu2022learning}} & L & 80.4 & 52.6 & 88.8 & 89.9 & 66.0 & 77.7 & 79.6 & 70.8 & 85.5 & 70.7 & 97.4 & 67.2 & 78.7 & 75.2 & 91.3 & 88.8 & 81.4 & -\\

    \multicolumn{1}{c|}{SPVCNN++~\cite{tang2020searching}} & L & 86.4 & 43.1 & 91.9 & \textbf{92.2} & 75.9 & 75.7 & 83.4 & 77.3 & 86.8 & \textbf{77.4} & 97.7 & 71.2 & 81.1 & 77.2 & 91.7 & 89.0 & 81.1 & 110 \\

    \multicolumn{1}{c|}{LidarMultiNet~\cite{ye2023lidarmultinet}} & L & 80.4 & 48.4 & 94.3 & 90.0 & 71.5 & \textbf{87.2} & 85.2 & 80.4 & 87.0 & 74.8 & 97.8 & 67.3 & 80.7 & 76.5 & 92.1 & 89.6 & 81.4 & -\\

    \multicolumn{1}{c|}{UDeerPep~\cite{lacrange}} & L & \textbf{85.5} & 55.5 & 90.5 & 91.6 & 72.2 & 85.6 & 81.4 & 76.3 & 87.3 & 74.0 & 97.7 & \textbf{70.2} & 81.1 & 74.4 & 92.7 & \textbf{90.2} & 81.8 & - \\

    \hline
    
    \multicolumn{1}{c|}{LaCRange~\cite{lacrange}} &  L+C & 78.0 & 32.6 & 88.3 & 84.5 & 63.9 & 81.5 & 75.6 & 72.5 & 64.7 & 68.0 & 96.6 & 65.9 & 78.6 & 75.0 & 90.4 & 88.3 & 75.3 & 50\\
    
    \multicolumn{1}{c|}{LIF-seg~\cite{zhao2023lif}} &  L+C & 58.1 & 36.3 & 86.7 & 84.3 & 60.0 & 79.7 & 80.3 & 77.8 & 83.2 & 68.7 & 97.2 & 68.2 & 77.0 & 74.5 & 91.0 & 89.0 & 75.7 & -\\
    
    \multicolumn{1}{c|}{PMF~\cite{zhuang2021perception}} &  L+C & 82.1 & 40.3 & 80.9 & 86.4 & 63.7 & 79.2 & 79.8 & 75.9 & 81.2 & 67.1 & 97.3 & 67.7 & 78.1 & 74.5 & 90.0 & 88.5 & 77.0 & $22^{\text{T}}$\\    

    \multicolumn{1}{c|}{CPFusion~\cite{lacrange}} &  L+C+R & 83.7 & 37.0 & 89.0 & 86.2 & 70.1 & 77.5 & 78.1 & 74.5 & 82.8 & 67.9 & 96.6 & 68.2 & 79.5 & 74.9 & 90.5 & 87.0 & 77.7 & - \\  
    
    \multicolumn{1}{c|}{2D3DNet~\cite{genova2021learning}} &  L+C & 83.0 & 59.4 & 88.0 & 85.1 & 63.7 & 84.4 & 82.0 & 76.0 & 84.8 & 71.9 & 96.9 & 67.4 & 79.8 & 76.0 & 92.1 & 89.2 & 80.0 & - \\

    \multicolumn{1}{c|}{MSeg3D~\cite{li2023mseg3d}} &  L+C & 83.1 & 42.5 & \textbf{94.9} & 92.0 & 67.1 & 78.6 & \textbf{85.7} & \textbf{80.5} & \textbf{87.5} & 77.3 & 97.7 & 69.8 & \textbf{81.2} & \textbf{77.8} & 92.4 & 90.1 & 81.4 & 445\\

    \hline
    \multicolumn{1}{c|}{CPGNet-LCF[Ours]} & L+C & 84.9 & \textbf{63.5} & 94.4 & \textbf{92.2} & \textbf{79.1} & 85.9 & 85.4 & 78.8 & 86.2 & 76.4 & \textbf{97.9} & 66.5 & 81.0 & 76.4 & \textbf{93.0} & 89.5 & \textbf{83.2} & \textbf{63/20$^{\text{T}}$} \\    

    \hline
    \end{tabular}}
  \label{tab:nus_test}
\end{table*}

\textbf{Implementation Details} of data process, model settings, and training details are as follows.
For a fair comparison with the previous SOTA method MSeg3D~\cite{li2023mseg3d}, the input image size is resized to $\left[640, 960\right]$ on the nuScenes dataset and $\left[360, 1280\right]$ on the SemanticKITTI dataset. The range of the point cloud is set as $\left[-51.2m, +51.2m\right]$ along the XY axes and $\left[-5.0m, +3.0m\right]$ along the Z axis on the nuScenes dataset. The range of the point cloud is set as $\left[-75.2m, +75.2m\right]$ along the XY axes and $\left[-4.0m, +2.0m\right]$ along the Z axis on the SemanticKITTI dataset. 
On both datasets, the BEV and RV branches of the LiDAR segmentation backbone accept a $512\times 512$ and a $64 \times 2048$ tensors, respectively. Besides, the model hyperparameters follows the CPGNet~\cite{li2022cpgnet} and the proposed CPGNet-LCF adopts a two-stage version of the CPGNet.


We use Adam~\cite{kingma2014adam} as the optimizer with a base learning rate of 0.001 and weight decay of 0.001. OneCycle~\cite{loshchilov2016sgdr} is used as the learning rate scheduler strategy with the annealing strategy of the cosine curve. 
Following the convention, the data augmentation strategies, containing random flipping along the XY axes, random global scale sampled from $\left[0.95, 1.05\right]$, random rotation around the Z axis, random Gaussian noise $\mathcal{N}(0,0.02)$ are used. The balanced coefficients of the loss terms are set as $\lambda_1=1, \lambda_2=1$.
\begin{table}[t]
    \caption{Angle noise range of each level.}
    \label{tab:nus_weak_degree}
    \centering
    \setlength\tabcolsep{3pt}
    \begin{tabular}{c|cccc}
    \hline
    \tabincell{c}{Level of \\ Weak Calibration} & \tabincell{c}{Level 0\\Well Calibration} & Level 1 & Level 2 & Level 3\\
    \hline
    \tabincell{c}{Angle Noise Range} & $\left[0\degree,0\degree\right]$ & $\left[-1\degree,1\degree\right]$ & $\left[-2\degree,2\degree\right]$ & $\left[-4\degree,4\degree\right]$\\
    \hline
    \end{tabular}
\end{table}
\textbf{Weak Calibration Benchmark} is established to evaluate the robustness of different models affected by the calibration errors. nuScenes is considered as well calibration dataset, which has been carefully processed and validated before and after data collection~\cite{caesar2020nuscenes}. The weak calibration with the corresponding disturbing matrices is illustrated in Sec.~\ref{sec:weak}. $r=0,1,2,3$ denotes the level of the weak calibration, as shown in Tab.~\ref{tab:nus_weak_degree}.
The weak calibration benchmark is built on the nuScenes validation set by randomly selecting the calibration angle noise from the predefined angle noise range of each level to form the disturbing matrices $(E_r^x, E_r^y, E_r^z)$. In the phases of weak calibration data augmentation and knowledge distillation, Level 3 is used during training. Note that Level 0 of the weak calibration refers to the well calibration.



\subsection{Results on Weak Calibration Benchmark}

\begin{table}[t]
    \caption{Comparison of different methods and training strategies on the different levels of weak calibration. DA and KD denote the weak calibration data augmentation and knowledge distillation, respectively.}
    \label{tab:nus_weak_calib}
    \centering
    \begin{tabular}{c|c|cccc}
    \hline    
     & & & & \\
    Methods & {\rotatebox{90}{LiDAR-only}} & {\rotatebox{90}{\tabincell{c}{Level 0}}} & {\rotatebox{90}{Level 1}} & {\rotatebox{90}{Level 2}} & {\rotatebox{90}{Level 3}} \\
    \hline
    MSeg3D & \multirow{2}{*}{72.0} & 80.7 & 79.4 & 77.2 & 73.1 \\
    MSeg3D + DA &  & 78.9 & 78.8 & 78.6 & 77.5 \\
    \hline
    CPGNet-LCF & \multirow{3}{*}{\textbf{79.0}} & 82.5 & 81.3 & 78.9 & 74.2 \\
    CPGNet-LCF + DA & & 81.9 & 81.9 & 81.5 & \textbf{80.1} \\
    CPGNet-LCF + KD & & \textbf{82.7} & \textbf{82.4} & \textbf{81.8} & \textbf{80.1} \\
    \hline
    \end{tabular}
\end{table}

Tab.~\ref{tab:nus_weak_calib} shows the results of our proposed weak calibration evaluation benchmark. LiDAR-only means that the model only has the input of LiDAR point cloud data. We compare the proposed CPGNet-LCF with the previous SOTA method MSeg3D~\cite{li2023mseg3d}, and the result shows that both MSeg3D and our proposed method CPGNet-LCF suffer from performance degradation as the calibration errors gradually increase. 
Even worse, at Level 3, the multi-modal CPGNet-LCF underperforms its LiDAR-only counterpart by a large performance drop(-8.3 mloU), which means that weak calibration greatly affects the performance of the multi-modal methods and deserves considerable exploration.

When the weak calibration data augmentation is considered, the performance of both MSeg3D~\cite{li2023mseg3d} and our CPGNet-LCF have a significant improvement, especially +4.4 mloU of MSeg3D and +5.9 mloU of CPGNet-LCF on the Level 3.
However, the performance on well calibration (Level 0) degrades due to the misleading training process caused by the weak calibration data augmentation.
Weak calibration knowledge distillation strategy can effectively solve this degradation problem under the well calibration, even improving the performance across all levels of the weak calibration. It demonstrates the effectiveness of the proposed training strategy in alleviating the weak calibration problems.

\begin{figure*}[t]
\centering
\subfloat[CPGNet LiDAR-Only]{
\label{cpg}
\includegraphics[width=0.23\linewidth]{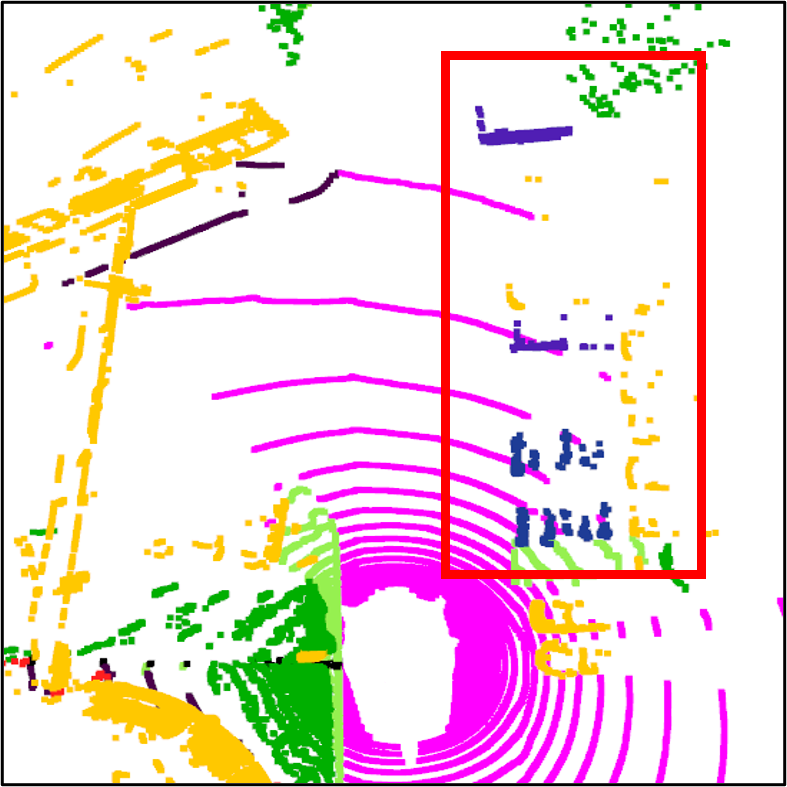}}
\subfloat[CPGNet-LCF (Well Calibration)]{
\label{lcf}
\includegraphics[width=0.23\linewidth]{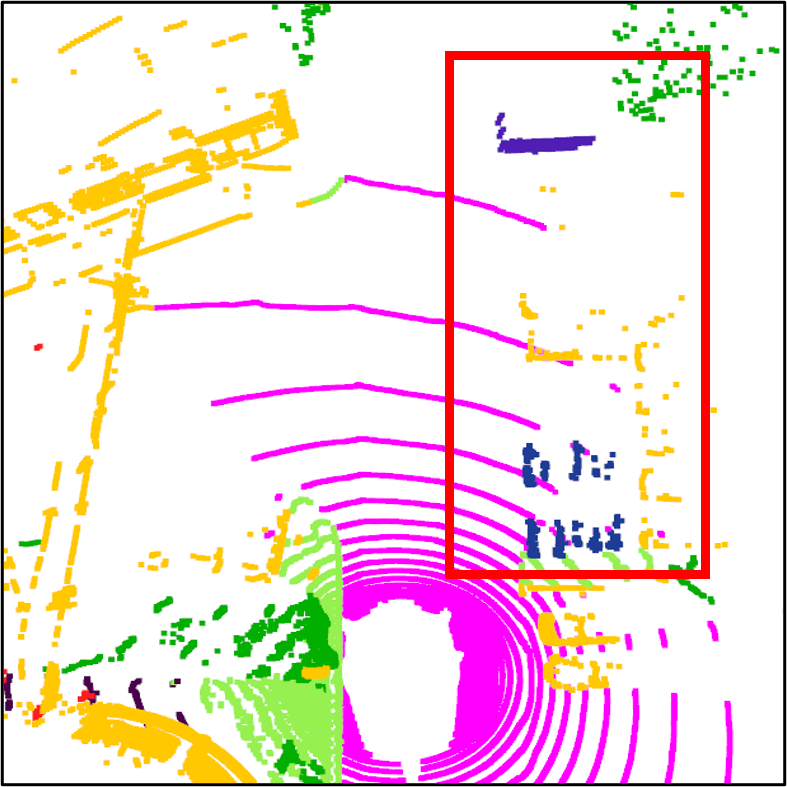}}
\subfloat[CPGNet-LCF (Level 3)]{
\label{lcf_l3}
\includegraphics[width=0.23\linewidth]{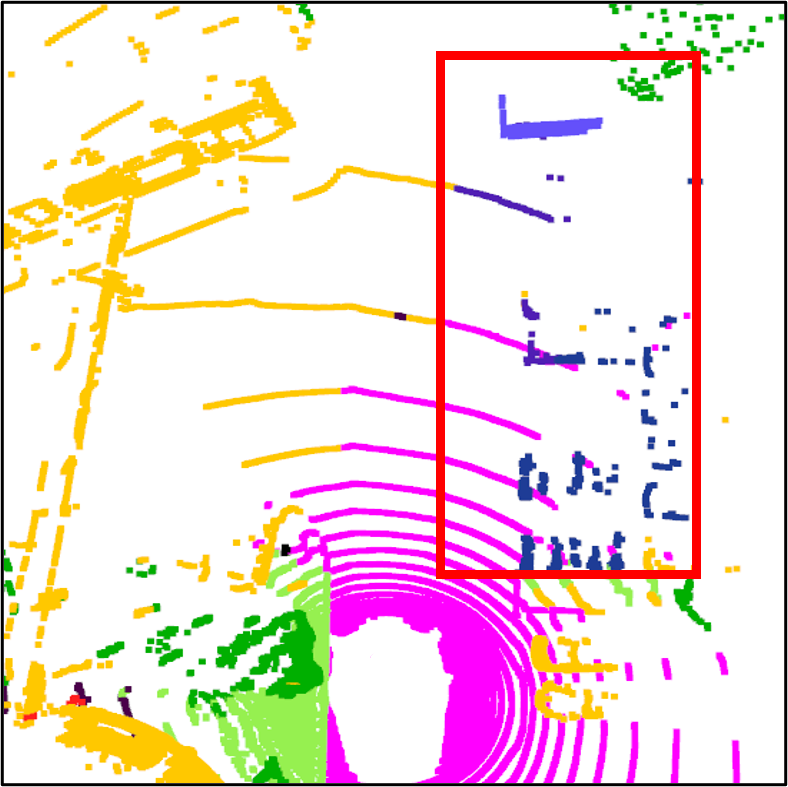}}
\subfloat[CPGNet-LCF+KD (Level 3)]{
\label{kd}
\includegraphics[width=0.23\linewidth]{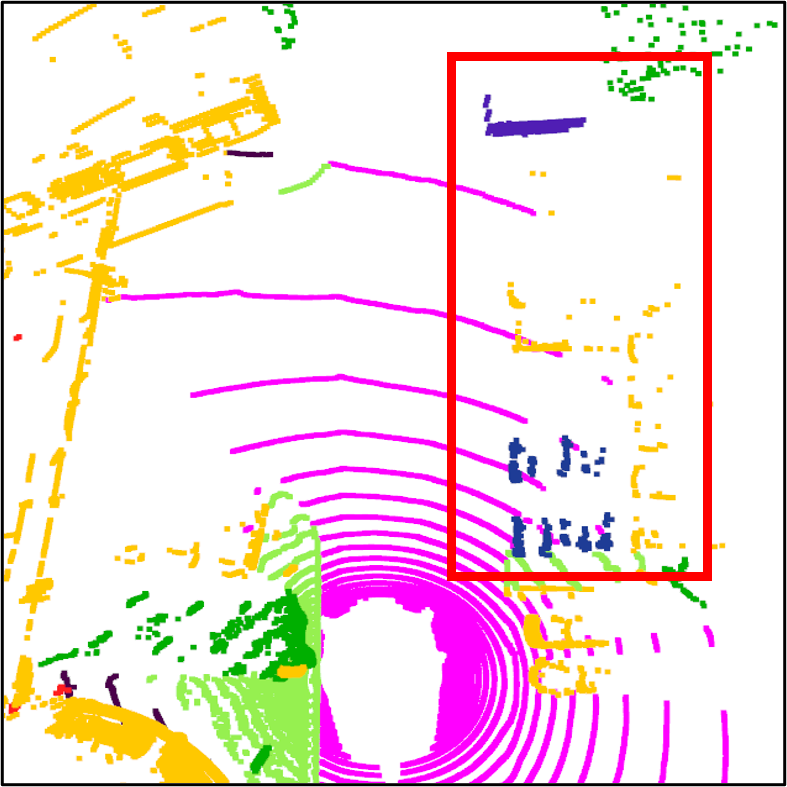}}

\caption{Visualization results of different methods on the nuScenes validation set.}
\label{fig:visual}
\end{figure*}

\begin{table}[t]
    \centering
    \caption{Comparison results on the SemanticKITTI validation dataset. }
    \label{tab:semkitti_test}
    \begin{tabular}{c|c|c}
    \hline
    Method & Modality & mIoU$^{1}$\\
    \hline
    SalsaNext~\cite{cortinhal2020salsanext} & L & 59.4 \\
    SPVNAS~\cite{tang2020searching} & L & 62.3 \\
    Cylinder3D~\cite{zhu2021cylindrical} & L & 64.9 \\
    PointPainting~\cite{vora2020pointpainting} & L+C & 54.5 \\
    PMF~\cite{zhuang2021perception} & L+C & 63.9\\
    MSeg3D~\cite{li2023mseg3d} & L+C& 66.7 \\
    \hline
    CPGNet-LCF[Ours] & L+C & \textbf{67.1} \\
    \hline
    \end{tabular}
\end{table}

\subsection{Comparisons With the State-of-the-Arts}

Comparison results of the nuScenes leaderboard and SemanticKITTI validation set are shown in Tab.~\ref{tab:nus_test} and Tab.~\ref{tab:semkitti_test}, respectively.
CPGNet-LCF outperforms all LiDAR-only and multi-modal methods on both accuracy and inference speed by a large margin on the nuScenes leaderboard. mloU$^1$ in Tab.~\ref{tab:semkitti_test} means the setting of PMF~\cite{zhuang2021perception}, which only evaluates the performance of the overlapped areas between LiDAR and cameras. The proposed CPGNet-LCF also achieves the best mloU of 67.1. These observations prove the superiority of the proposed framework.
In addition, our model also has significant advantages in deployment and speed that it runs 63\,ms per frame with PyTorch and 20\,ms per frame with TensorRT TF16 on a single NVIDIA Tesla V100 GPU.

In Tab.\ref{tab:image}, we also evaluate the impacts of different image backbones. In the proposed CPGNet-LCF framework, the STDC1~\cite{fan2021rethinking} outperforms the HRNet-w18~\cite{wang2020deep} on both accuracy and inference speed, demonstrating the superiority of the STDC1. Besides, with the same image backbone, HRNet-w18, the proposed CPGNet-LCF outperforms the previous SOTA MSeg3D~\cite{li2023mseg3d} on both accuracy and inference speed since the MSeg3D adopts the time-consuming 3D sparse convolution as the basic operator.
\subsection{Visualization}
Visualization results on the nuScenes validation set are shown in Fig.~\ref{fig:visual}.
It can be seen that the LiDAR-only CPGNet (a) misclassifies several LiDAR points of \emph{Manmade} into \emph{Truck}, while the multi-modal CPGNet-LCF (b) can make correct predictions due to the rich texture information from cameras.
However, if the calibration matrices are disturbed on Level 3 (c), the predictions inferred by the same model (b) are even worse than those of the LiDAR-only CPGNet (a) since the misalignment between the LiDAR and cameras will confuse the objects with their neighbors. Surprisingly, this problem can be solved (d) by the proposed weak calibration knowledge distillation.

\begin{table}[t]
    \caption{Impacts of the image backbone on the MSeg3D and CPGNet-LCF.}
    \label{tab:image}
    \centering
    \begin{tabular}{c|c|cc}
    \hline
    Method & Image Backbone & mIoU & Latency(ms) \\
    \hline
    MSeg3D~\cite{li2023mseg3d} & HRNet-w18~\cite{wang2020deep} & 80.0 &  265\\
    \hline
    \multirow{3}{*}{CPGNet-LCF} & $\times$ & 79.0 & \textbf{43} \\
    & HRNet-w18~\cite{wang2020deep} & 81.4 & 96 \\
    & STDC1~\cite{fan2021rethinking} & \textbf{82.3} & 63 \\
    \hline
    \end{tabular}
\end{table}

\section{CONCLUSIONS}

In this paper, we first point out that easy-deployed, real-time, and robust against weak calibration are significant for real-world autonomous driving systems, which are under-explored in existing multi-modal methods. The proposed LiDAR and camera fusion model, dubbed CPGNet-LCF, is an extension of the easy-deployed and real-time CPGNet. Experimental results on the two public datasets demonstrate that the proposed CPGNet-LCF achieves the new SOTA results. Meanwhile, it is easily deployed on the TensorRT TF16 mode and runs 20\,ms on a single Tesla V100 GPU to meet the real-time inference requirement. Besides, the results of the multi-modal methods degrade under the weak calibration, even worse than the LiDAR-only counterparts, which is effectively alleviated by the proposed weak calibration knowledge distillation strategy. The weak calibration deserves more explorations in the future works.

\addtolength{\textheight}{-0cm}   
\clearpage
\bibliographystyle{IEEEtran}
\bibliography{IEEEabrv,sample}

\end{document}